\pdfoutput=1

\documentclass[11pt]{article}

\usepackage{acl}

\usepackage{times}
\usepackage{latexsym}

\usepackage[T1]{fontenc}

\usepackage[utf8]{inputenc}

\usepackage{microtype}

\usepackage{inconsolata}

\usepackage[ruled, linesnumbered]{algorithm2e}
\usepackage{graphicx}
\usepackage{amssymb}
\usepackage{amsmath}
\usepackage[ruled, linesnumbered]{algorithm2e}
\usepackage{booktabs}
\usepackage{amsfonts}
\usepackage{multirow}
\usepackage{makecell}
\usepackage{subfigure}
\usepackage{color}
\usepackage{bm}
\usepackage{algorithmic}

\usepackage{enumitem}

\title{Optimal Transport Guided Correlation Assignment \\ for Multimodal Entity Linking}

\author{
  Zefeng Zhang\thanks{\ These authors contribute equally to this work.}, Jiawei Sheng\footnotemark[1], Chuang Zhang\thanks{\ Corresponding Authors.},  \\
  \textbf{Yunzhi Liang}, \textbf{Wenyuan Zhang}, \textbf{Siqi Wang}, \textbf{Tingwen Liu}\\
  Institute of Information Engineering, Chinese Academy of Sciences. Beijing, China \\
  School of Cyber Security, University of Chinese Academy of Sciences. Beijing, China \\
  \texttt{\{zhangzefeng,shengjiawei,zhangchuang\}@iie.ac.cn}\\
  \texttt{\{liangyunzhi,zhangwenyuan,wangsiqi2022,liutingwen\}@iie.ac.cn}\\
}

\begin{document}
\maketitle
\begin{abstract}
Multimodal Entity Linking (MEL) aims to link ambiguous mentions in multimodal contexts to entities in a multimodal knowledge graph.
A pivotal challenge is to fully leverage multi-element correlations between mentions and entities to bridge modality gap and enable fine-grained semantic matching.
Existing methods attempt several local correlative mechanisms, relying heavily on the automatically learned attention weights, which may over-concentrate on partial correlations.
To mitigate this issue, we formulate the correlation assignment problem as an optimal transport (OT) problem, and propose a novel MEL framework, namely OT-MEL, with OT-guided correlation assignment.  
Thereby, we exploit the correlation between multimodal features to enhance multimodal fusion, and the correlation between mentions and entities to enhance fine-grained matching.
To accelerate model prediction, we further leverage knowledge distillation to transfer OT assignment knowledge to attention mechanism.
Experimental results show that our model significantly outperforms previous state-of-the-art baselines and confirm the effectiveness of the OT-guided correlation assignment.\footnote{Source code of this paper could be obtained from https://github.com/zhangzef/OT-MEL}
\end{abstract}

\section{Introduction} \label{sec:intro}

\begin{figure}[!t]
    \centering
    \includegraphics[width=0.48\textwidth]{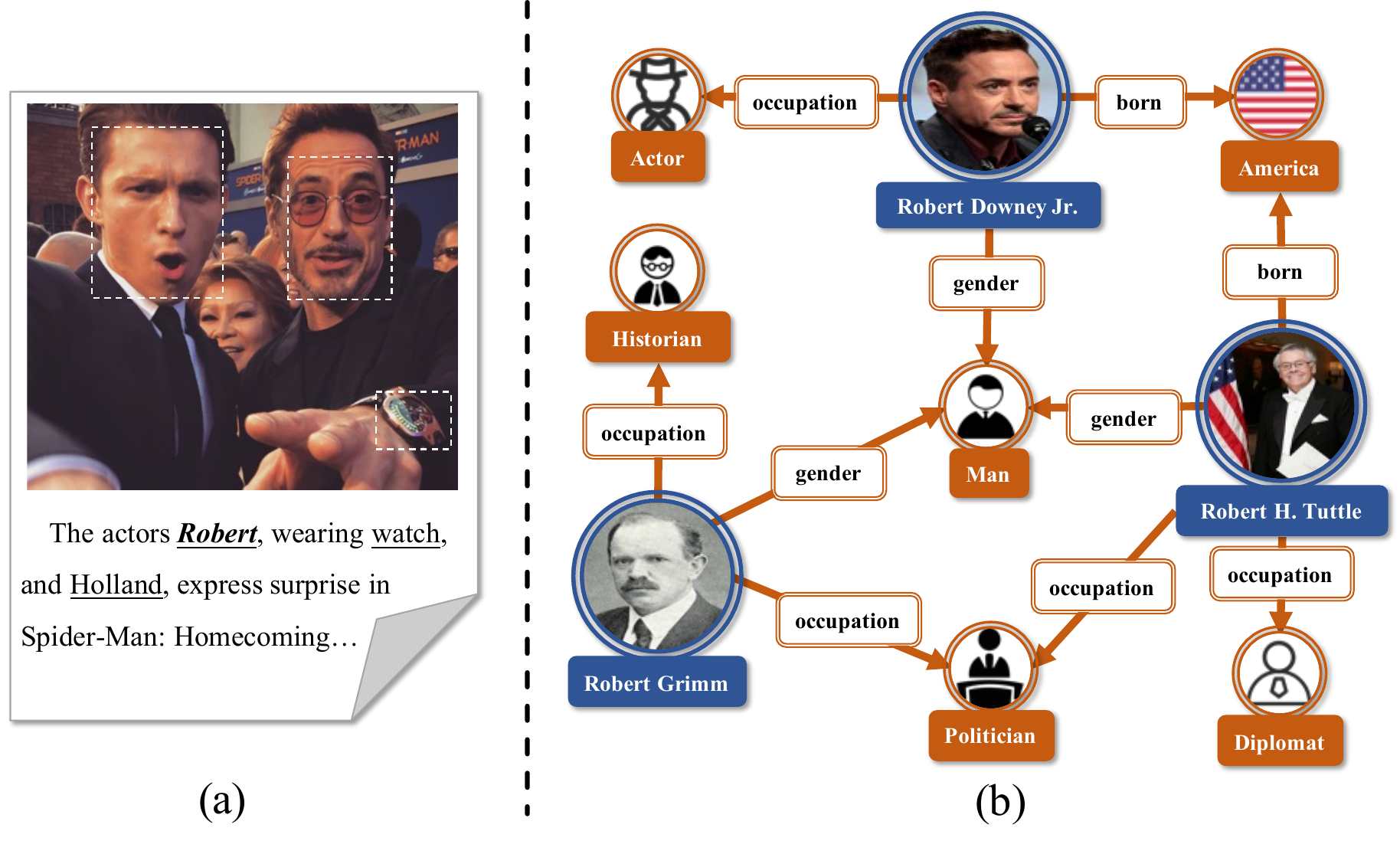}
    \caption{Examples of MEL: (a) the given mention ``Robert'' in a multimodal content, and (b) the candidate entities in the multimodal KG.}
    \label{fig:Examples}
\end{figure}

Entity Linking (EL) is an important yet challenging task in knowledge acquisition, which can facilitate applications such as information retrieval~\cite{liu-etal-2018-entity}, question answering~\cite{xiong-etal-2019-improving} and dialogue systems~\cite{jiang-etal-2023-cognitive}.
In recent years, with the rapid increase of multimodal content, Multimodal Entity Linking (MEL) has been proposed, which aims to link ambiguous \textit{mentions} in multimodal contexts to \textit{entities} in the multimodal knowledge graph (KG).
As shown in Figure~\ref{fig:Examples}, given a mention {\small \texttt{Robert}} in texts with related images, the MEL model requires the mention to be corresponded to the entity {\small  \texttt{Robert\_Downey}} in the multimodal KG.

Since the mention can be ambiguous with limited textual information, it is crucial to enrich the text with its related visual images.
To utilize multimodal information for MEL, a pivotal challenge is to leverage \textit{multi-element correlations} within entity and mention contexts:
1) \textit{the multi-element correlations between modalities}, where the textual tokens and visual patches may have correlative correspondence.
As shown in Figure~\ref{fig:Examples}, the token {\small \texttt{Robert}} and {\small \texttt{Holland}} correspond to different local areas of human faces in the image.
These correlations serve as message-passing pivots between the text and image, which can be valuable to bridge the semantic gap of modalities~\cite{luo2023multi}.
2) \textit{the multi-element correlations between mentions and entities}, where the context of mentions and entities can have multiple correlative cues for semantic matching.
As shown in Figure~\ref{fig:Examples}, the mentions and entity both involve token {\small \texttt{Robert}}, and both the mention and the entity involve the word {\small \texttt{Actor}}.
These correlations depict similarities between the mention and entity, providing explicit evidence for EL.

To facilitate these multi-element correlations, existing studies propose sophisticated correlative mechanisms, such as feature concatenation~\cite{adjali2020multimodal}, attention-based modules~\cite{moon2018multimodal,wang2022multimodal}, hierarchical gates~\cite{luo2023multi,wang2022multimodal} and graph-based aggregation~\cite{xing2023drin}.
However, these models learn the correlations of elements locally, and rely heavily on the automatically learned correlation assignments.
This design may over-concentrate on part of elements, dominating the correlations without global consideration.
For example, all entities in Figure~\ref{fig:Examples} have surface name {\small \texttt{Robert}} the same to mention {\small \texttt{Robert}}, which may attract the correlation assignment and conceal crucial correlations of token {\small \texttt{Actor}} in texts and the visual patch of human face in images.


To mitigate the above issues, we formulate the multi-element correlation assignment as an \textit{optimal transport (OT)} problem.
Taking the multimodal correlation as an example, given the cost of each pair of textual tokens and visual patches, the OT problem aims to find an optimal transport plan to transfer the token to a patch with assigned weights, such that the total cost is minimal.
In contrast to the automatically learned attention map, the OT plan globally considers the correlation between source and target elements, surpassing the degraded dominant correspondence~\cite{DBLP:conf/cvpr/LiuZYY20}.

Following the above idea, we propose a novel \underline{MEL} framework with \underline{OT} guided correlation assignment, termed \textbf{OT-MEL}. 
Specifically, we first adopt multimodal feature encoders to represent textual tokens and visual patches for both mentions and entities.
Afterward, we propose an OT-based correlation assignment method between textual tokens and visual patches, with which we further integrate multimodal features and bridge the semantic gap of modalities.
Additionally, we also build unimodal correlations between mentions and entities with OT-based correlation assignment.
By measuring the similarity of the multimodal and unimodal features between mentions and entities, the model derives the overall matching score. 
To enhance efficiency, we further propose distillation between OT assignment to attention map, such that we can use attention to approximate OT plan.
The contributions of this paper are three-fold:
\begin{itemize}[leftmargin=*]
\item We introduce optimal transport to capture complicated correlations in MEL.
To our knowledge, we are the first to guide MEL with OT.
\item We propose OT-MEL, a novel framework to consider the multimodal and unimodal correlations, which can be further extended by knowledge distillation from OT plan to attention mechanism.
\item Empirical studies indicate that our model outperforms previous state-of-the-art methods on three widely-used benchmarks and extensive analysis confirm the effectiveness of our model.
\end{itemize}

\section{Preliminary}

\subsection{Task Formulation}
The task of MEL aims to correspond a given mention from text within a multimodal context to its true entity in a multimodal KG.
Generally, the KG involves entities $e\in\mathcal{E}$, where each entity $e\triangleq\{e^n, e^v, e^a\}$ can be associated with entity name $e^n$, visual image $e^v$, and a textual attribute set $e^a$ \footnote{Following most existing MEL studies\cite{wang2022multimodal,luo2023multi}, we also treat the attributes as textual information of the entities due to the large-scale KG consideration.}.
Besides, the given mention from texts can be summarized as $m \triangleq \{m^n, m^c, m^v\}$, which is associated with mention word $m^n$, context sentence $m^c$, and visual image $m^v$, respectively. 

To achieve this task, for each given mention, the model requires to identify the true entity from entity candidates, which can be formulated as: 
\begin{equation}\label{eq:MELTarget}
\theta^{*} = \underset{\theta}{\arg\max}\sum_{(m, e)\in \mathcal{D}} \log p_{\theta}(e|m,\mathcal{E}),
\end{equation}
where $\theta^{*}$ denotes the final model.
Following previous studies~\cite{luo2023multi}, we use all entities in KG as the candidates.

\subsection{Optimal Transport}

\begin{figure*}[!ht]
    \centering
    \includegraphics[width=1\textwidth]{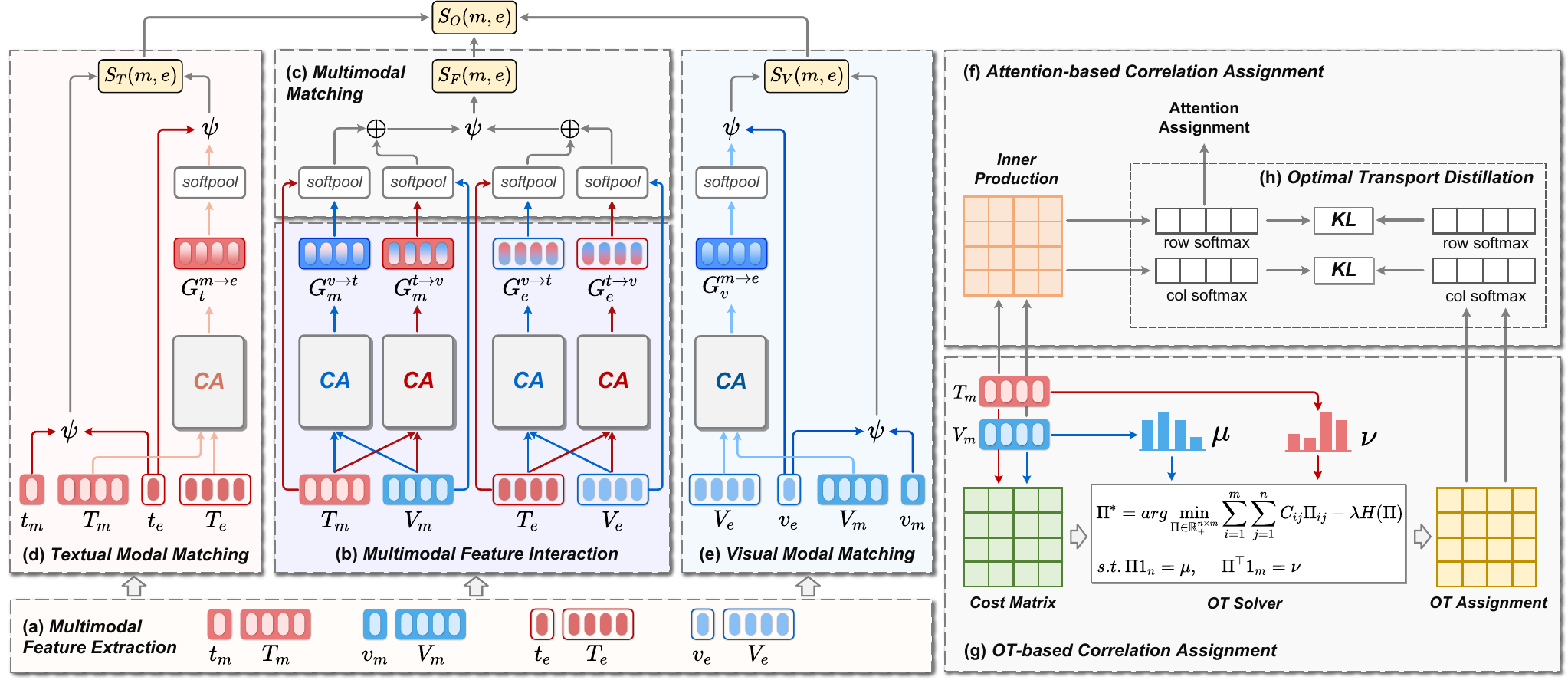}
    \caption{The overall OT-MEL framework. CA denotes the correlation assignment modules with attention or OT.}
    \label{fig:framework}
    \vspace{-1em}
\end{figure*}

Optimal Transport (OT) aims to find a transport plan with minimal total cost, which transports the density distribution of a group of elements to that of another group, given the cost of each element pair~\cite{villani2009optimal}.
Formally, given element $x_i$ in one group, $y_j$ in another group and $\bm{C}_{ij}$ the cost between them, the problem can be formulated as the following linear programming problem:
\begin{equation}\label{eq:Kantorovich-discrete2}
\begin{aligned}
    & \bm{\Pi}^\star = \underset{\bm{\Pi} \in \mathbb{R}_{+}^{n \times m}}{\mathrm{\arg \min}} \sum_{i=1}^{m} \sum_{j=1}^{n} \bm{C}_{ij} \bm{\Pi}_{ij}, \\
    & \mathrm{s.t.} \ \bm{\Pi} \mathbf{1}_n = \bm{\mu}, \quad \bm{\Pi}^{\top} \mathbf{1}_m = \bm{\nu},
\end{aligned}
\end{equation}
where $\bm{\Pi}^*$ is the optimal transport plan, where $\bm{\Pi}_{ij}$ is the amount of information from $x_i$ to $y_j$.
$\mathbf{1}_n$ and $\mathbf{1}_m$ are all-ones vectors of dimension $n$ and $m$, respectively.
In this way, the transport plan remains the multi-element correlations between two groups with appropriate weight assignments. 

To accelerate the calculation, ~\citet{cuturi2013sinkhorn} proposes to smooth the problem with the entropy regularization term as follows:
\begin{equation}
\begin{aligned}
    & \bm{\Pi}^\star = \underset{\bm{\Pi} \in \mathbb{R}_{+}^{n \times m}}{\mathrm{\arg \min}} \sum_{i=1}^{m} \sum_{j=1}^{n} \bm{C}_{ij} \bm{\Pi}_{ij} - \lambda H(\bm{\Pi}), \\
    & \mathrm{s.t.} \ \bm{\Pi} \mathbf{1}_n = \bm{\mu}, \quad \bm{\Pi}^{\top} \mathbf{1}_m = \bm{\nu},
\end{aligned}
\end{equation}
where $H(\bm{\Pi})$ is the entropy of $\bm{\Pi}$, $\lambda$ is regularization coefficient. 
Based upon, the OT problem can be efficiently solved by Sinkhorn-Knopp algorithm, detailed in Algorithm~\ref{Sinkhorn}.

\section{Method}

In this section, we introduce our OT-MEL model, which consists of three modules: 1) multimodal feature encoder, 2) multimodal feature interaction, and 3) multimodal entity linking. 
The architecture of OT-MEL is shown in Figure~\ref{fig:framework}.

\subsection{Multimodal Feature Extraction}

Given the mention $m$ and entity $e$ with their text and image, we are going to generate their initial embeddings. 
To enrich fundamental correlations between multimodal features, following \citet{luo2023multi}, we adopt contrastive language-image pre-training architecture (CLIP)~\cite{radford2021learning} as the multimodal encoder, which contains a pre-trained BERT~\cite{devlin2018bert} for textual embedding and a pre-trained ViT~\cite{dosovitskiy2020image} for visual embedding.
For entities and mentions we use a parameter shared encoder to ensure consistency in the representation space.

\paragraph{Textual Feature Extraction}
We employ BERT to extract textual features.
For mention $m$, we concatenate the word of mention $m_n$ with its context sentence $m_c$, and then feed it into BERT as:
\begin{equation}
    \bm{T}_{m} = \mathrm{BERT}([CLS] m_n [SEP] m_c [SEP]),
\end{equation}
where $\bm{T}_{m}\in \mathbb{R}^{L_{mt} \times d}$ is the mention textual embeddings.
On the other hand, for entity $e$, we concatenate the entity name $e_n$ with its attributes\footnote{Following \citet{luo2023multi}, we also treat different attributes as texts separated by period.} $e_a$ as \texttt{\small $[CLS] e_n [SEP] e_a [SEP]$}, and use the same way to obtain the entity textual embeddings $\bm{T}_{e} \in \mathbb{R}^{L_{et} \times d}$.

\paragraph{Visual Feature Extraction}
We employ the pre-trained ViT to extract visual features.
To process the mention image into token sequence, we reshape the image $\bm{m}^v \in \mathbb{R}^{H \times W \times C}$ into a sequence of flattened 2D patches as $\bm{m}^{v}_{p}\in \mathbb{R}^{P \times P \times C}$ with special token \texttt{\small [CLS]}, and feed them into ViT as:
\begin{equation}
    \bm{V}_{m} = \mathrm{ViT}([\bm{m}^v_{\texttt{[CLS]}}, \bm{m}^v_{1}, ..., \bm{m}^v_{l_v-1}]),
\end{equation}
where $ \bm{T}_{m}\in \mathbb{R}^{L_{mv} \times d}$ is the mention visual embeddings, and $l_v = HW/P^2$ is the resulted patch number.
Similarly, we reshape the image of entity into patches with the same patch size, and obtain the entity visual embeddings $\bm{V}_{e} \in \mathbb{R}^{L_{ev} \times d}$.

\subsection{Multimodal Feature Interaction}
As claimed in the introduction, the text and image can have multi-element correlations served as message passing pivots.
Therefore, we assign distinct weights for multimodal feature interaction.
Take the mention-side texts and images as examples, we first introduce an intuitive attention-based assignment, and thereupon propose our OT-based correlation assignment method.

\paragraph{Attention-based Correlation Assignment}
To interact multimodal features with distinct weights, an idea is to employ attention mechanism~\cite{transformer} between the text and image.
Given the mention $m$ with its textual embeddings $\bm{T}_{m}$ and visual embeddings $\bm{V}_{m}$, we establish the image-to-text correlation assignments as follows:
\begin{align}\label{eq:qkv}
    &\bm{Q} = \bm{T}_{m}\bm{W}_{q}, 
    \bm{K} = \bm{V}_{m}\bm{W}_{k}, 
    \bm{H} = \bm{V}_{m}\bm{W}_{v}, \\
    &\bm{A}^{v\rightarrow t} = \mathrm{softmax}(\bm{S}), \bm{S} = \bm{Q}\bm{K}^{\top}/\sqrt{d} \label{eq:attention_assignment} \\ 
    &\bm{G}_{m}^{v\rightarrow t} = \bm{A}^{v\rightarrow t}\bm{H}
\end{align}
where $\bm{W}_{q}$, $\bm{W}_{k}$, $\bm{W}_{v}$ are learnable parameters with $\mathbb{R}^{d\times d}$.
Here $\bm{A}_{ij}^{v\rightarrow t}\in \mathbb{R}^{L_{mt} \times L_{mv}}$ is the assigned  correlation weight from visual patch $\bm{V}_{m,j}$ to textual token $\bm{T}_{m,i}$.
As claimed, the attention-based assignment focuses on local similarity of element-pairs without global consideration.

\paragraph{OT-based Correlation Assignment}

To globally consider the correlations, we propose OT-based correlation assignment method.
Practically, we formulate the image-to-text correlation assignment as an OT problem, where we hope the textual tokens and visual patches has appropriate correlation with minimal total transport cost.
Therefore, we project original textual feature $\bm{T}_{m}$ and visual feature $\bm{V}_{m}$ into an assignment feature space, and define the transport cost from patches to tokens with a measurement $\phi$ as follows:
\begin{align}
    &\bm{Q} = \bm{T}_{m}\bm{W}_{q}, 
    \bm{K} = \bm{V}_{m}\bm{W}_{k}, 
    \bm{H} = \bm{V}_{m}\bm{W}_{h}, \\
    &\bm{C}_{ij} = \phi(\bm{Q}_{i}, \bm{K}_{j}) \triangleq \frac{1}{2}\ [1- \cos(\bm{Q}_{i}, \bm{K}_{j})], \label{eq:cost_matrix}
\end{align}
where we achieve $\phi$ with cosine similarity such that $\phi\in [0,1]$. 
In this way, the token-patch pairs with higher similarity would has lower transport cost.
Based upon, we treat each element of tokens and patches equally as uniform distributions, and solve the correlation assignment problem by Sinkhorn-Knopp algorithm~\cite{cuturi2013sinkhorn}:
\begin{align}
    &\bm{A}^{v\rightarrow t}=\bm{\Pi^{\star}}\!\! =\! \mathrm{Sinkhorn}\text{-}\mathrm{Knopp}(\bm{C}, \bm{\mu}, \bm{\nu}), \label{eq:ot_assignment}\\
    &\bm{G}_{m}^{v\rightarrow t} = \bm{A}^{v\rightarrow t}\bm{H}    
\end{align}
where $\bm{\mu}= 1/L_{mt}\mathbf{1}_{mt}$ and $\bm{\nu}= 1/L_{mv}\mathbf{1}_{mv}$ are the probabilistic simplexes of textual tokens and visual patches, respectively.
In this way, we establish the image-to-text correlations with OT assignment in Eq.~(\ref{eq:ot_assignment}), which ensures the overall transport cost to be minimal.
Actually, this design treats all tokens and patches equally, and ensures all token-patch pairs to be appropriately transported according to the similarity-related cost, thus avoiding the suboptimal assignment dominated by partial elements.
In summary, the above procedure holds for both mentions and entities, and all generated features are termed as $\bm{G}_{m}^{v\rightarrow t}$, $\bm{G}_{m}^{t\rightarrow v}$, $\bm{G}_{e}^{t\rightarrow v}$ and $\bm{G}_{e}^{v\rightarrow t}$.

\subsection{Multimodal Entity Linking}

With the interacted multimodal features, we conduct MEL by matching the mentions and entities.
To enhance the interaction between unimodal features of mentions and entities, we also conduct OT-based correlation assignment on them.

\paragraph{Multimodal Feature Matching}

To fully consider the multimodal information, we integrate the interacted features to match mentions and entities, respectively.
Take mention $m$ as an example, we integrate the image-to-text representation with its textual embedding $\bm{T}_{m}$ and the transported embedding $\bm{G}_{m}^{v\rightarrow t}$ from its image.
Generally, we can employ any pooling method to aggregate these embeddings as $\mathcal{G}=\{\bm{T}_{m},\bm{G}_{m}^{v\rightarrow t}\}$. 
Here we introduce a simple yet effective method, softpool~\cite{stergiou2021refining}, which is a soft version of max-pooling: 
\begin{align}\label{eq:softpool1}
    &\bar{\bm{g}}_{m}^{t}=\mathrm{softpool}(\mathcal{G}) \triangleq \sum_{\bm{g}_{i}\in \mathcal{G}} \bm{w}_{i}\odot\bm{g}_i, \\
    &\bm{w}_{i} = \frac{\exp(\bm{g}_i)}{\sum_{\bm{g}_{j}\in \mathcal{G}}\exp(\bm{g}_j)}
\end{align}
where $\bar{\bm{g}}_{m}^{t}\in \mathbb{R}^{d}$ is the generated textual representation, $\odot$ is Hadamard production.
In this way, it emphasizes the most activated features in the generated multimodal representation.

Based upon, we obtain the text-to-image representations of mention $m$, and those representations of entity $e$ as follows:
\begin{align}\label{eq:softpool2}
    \bar{\bm{g}}_{m}^{v}&=\mathrm{softpool}(\{\bm{V}_{m},\bm{G}_{m}^{t\rightarrow v}\})\\ 
    \bar{\bm{g}}_{e}^{t}&=\mathrm{softpool}(\{\bm{T}_{e},\bm{G}_{e}^{v\rightarrow t}\})\\
    \bar{\bm{g}}_{e}^{v}&=\mathrm{softpool}(\{\bm{V}_{e},\bm{G}_{e}^{t\rightarrow v}\})
\end{align}
Finally, we measure the matching scores between mentions and entities, considering both of their textual and visual representations:
\begin{align}\label{eq:softpool3}
    S_{\mathrm{F}}(m,e) = \psi ({\bar{\bm{g}}}_{m}^{t} \oplus {\bar{\bm{g}}}_{m}^{v},\bar{\bm{g}}_{e}^{t} \oplus \bar{\bm{g}}_{e}^{v})
\end{align}
where $\psi$ is the similarity measurement, and we achieve it by vector inner production.
$S_{\mathrm{F}}(m,e)$ is the fused multimodal feature matching score.

\paragraph{Unimodal Feature Matching}
Considering the unimodal correlations between mentions and entities, we design unimodal feature matching module with OT-based correlation assignment. 
Specifically, take the text-modal as an example, we capture the correlations between mention textual feature $\bm{T}_{m}$ to entity $\bm{T}_{e}$.
We also derive the cost matrix with cosine similarity between them, and obtain the textual modal correlated mention representation:
\begin{align}
    &\bm{Q} = \bm{T}_{e}\bm{W}_{q}^{'}, 
    \bm{K} = \bm{T}_{m}\bm{W}_{k}^{'}, 
    \bm{H} = \bm{T}_{m}\bm{W}_{v}^{'}, \\
    &\bm{C}_{ij} = \phi(\bm{Q}_{i}, \bm{K}_{j}) \triangleq \frac{1}{2}\ [1- \cos(\bm{Q}_{i}, \bm{K}_{j})],  \\
    &\bm{A}^{m\rightarrow e}\!\!=\!\bm{\Pi^{\star}}\!\! = \mathrm{Sinkhorn}\text{-}\mathrm{Knopp}(\bm{C}, \bm{\mu}, \bm{\nu}),\\
    &\bm{G}_{t}^{m\rightarrow e} = \bm{A}^{m\rightarrow e}\bm{H}
\end{align}
where $\bm{\mu}= 1/L_{mt}\mathbf{1}_{mt}$ and $\bm{\nu}= 1/L_{et}\mathbf{1}_{et}$ are the probabilistic simplexes.
Based upon, we aggregate the textual representations for the mention, and achieve the unimodal feature matching between mention and entity by:
\begin{align}\label{eq:softpool4}
    &\bar{\bm{g}}_{m}=\mathrm{softpool}(\{\bm{G}_{t}^{m\rightarrow e}\}) \\
    &S_{\mathrm{T}}(m,e) = \frac{1}{2}[\psi (\bar{\bm{g}}_{m}, \bm{t}_{e}) + \psi ({\bm{t}}_{m}, \bm{t}_{e})]
\end{align}
where we also use inner production to achieve $\psi$.
Here, $\bm{t}_{m}$ and $\bm{t}_{e}$ are the output embedding of token \texttt{\small [CLS]}, remaining the original summarized feature of texts.
This precedure also holds for visual modal feature matching with matching score $S_{\mathrm{V}}(m,e)$.

\paragraph{Overall Matching Score}
To consider different aspects in feature matching, we collect all matching scores and obtain the overall matching score as:
\begin{equation}
\begin{aligned}\label{eq:overall_score}
    & S_{\mathrm{O}}(m,e) = \frac{1}{3}\sum_{\mathrm{X}\in \{\mathrm{F},\mathrm{T},\mathrm{V}\}}  S_{\mathrm{X}}(m,e)
\end{aligned}
\end{equation}
Thereby, we design contrastive training objective on the overall matching score as follows:
\begin{equation}
\begin{aligned}\label{eq:contrastive loss1}
    \mathcal{L}_{\mathrm{O}} =  -\log\frac{\exp(S_{\mathrm{O}}(m, e))}{\sum_{e'\in \mathcal{E}} \exp (S_{\mathrm{O}}(m, e'))},
\end{aligned}
\end{equation}
where $\mathcal{E}$ is all candidate entities in the KG.
For training, we adopt all other entity $e'$ except the true entity $e$ in the batch, which serve as the in-batch negative samples.
To enhance the consistency in different matching aspects, following~\citet{luo2023multi}, we also apply contrastive training objective on different matching scores:
\begin{equation}
\begin{aligned}\label{eq:contrastive loss2}
    \mathcal{L}_{\mathrm{X}} &=  -\log\frac{\exp(S_{\mathrm{X}}(m, e))}{\sum_{e'\in \mathcal{E}} \exp (S_{\mathrm{X}}(m, e'))},\\
    \mathcal{L}_{}&= \mathcal{L}_O + \sum_{\mathrm{X}\in \{\mathrm{F},\mathrm{T},\mathrm{V}\}}\mathcal{L}_{\mathrm{X}}
\end{aligned}
\end{equation}
where $\mathrm{X}\in \{\mathrm{F},\mathrm{T},\mathrm{V}\}$ is the original multimodal and unimodal matching scores.
$\mathcal{L}$ is the overall loss of our model with the full OT-based version.

\subsection{Optimal Transport Distillation}
To improve the efficiency of OT-based correlation assignments, 
we further propose an efficient version with KD, which transfers knowledge from OT-based correlation assignment to attention-based correlation assignment.
Generally, we can use mean squared error (MSE) loss to assimilate the assignment $\bm{A}^{v\rightarrow t}$ in Eq.~(\ref{eq:attention_assignment}) and Eq.~(\ref{eq:ot_assignment}).
To better consider both the source and target aspects in OT assignment, we design Kullback-Leibler (KL) divergence to fully transfer knowledge as follows:
\begin{equation}\label{eq:kl}
\begin{aligned}
    & \mathrm{KD}(\bm{A}_{\mathrm{OT}}^{v\rightarrow t}, \bm{A}_{\mathrm{ATT}}^{v\rightarrow t}) \\
    &\triangleq \frac{1}{2}[\begin{matrix}\sum_{j} \end{matrix}\mathrm{KL}(\mathrm{softmax}(\bm{\Pi}^{\star}_{i*}), \mathrm{softmax}(\bm{S}_{i*})) \\ 
    &+ \begin{matrix}\sum_{j} \end{matrix} \mathrm{KL}(\mathrm{softmax}(\bm{\Pi}^{\star}_{*j}), \mathrm{softmax}(\bm{S}_{*j}))]
\end{aligned}
\end{equation}
Here, we conduct knowledge distillation on all OT-based correlation assignments, including $\bm{A}^{m, v\rightarrow t}$, $\bm{A}^{m, t\rightarrow v}$,$\bm{A}^{e, v\rightarrow t}$, $\bm{A}^{e, t\rightarrow v}$ for multimodal feature interaction, and  $\bm{A}^{m\rightarrow e}$, $\bm{A}^{e\rightarrow m}$ for unimodal feature matching.
All these knowledge distillation constraints constitute the overall KD loss $\mathcal{L}_{\mathrm{KD}}$.
Alternatively, the overall training loss is:
\begin{equation}\label{eq:optimization loss}
    \mathcal{J}= \mathcal{L} + \mathcal{L}_{\mathrm{KD}}
\end{equation}
Here, $\mathcal{J}$ is the overall loss with knowledge distillation, which obtains an efficient KD-version model.

\begin{table*}[t]
    \centering\footnotesize
    \setlength{\tabcolsep}{5pt}
    \begin{tabular*}{\textwidth}{@{\extracolsep{\fill}}@{}l|cccc|cccc|cccc@{}}
    \toprule
    \multirow{2}{*}{Model} &  \multicolumn{4}{c|}{RichpediaMEL} &
    \multicolumn{4}{c|}{WikiMEL} & \multicolumn{4}{c}{WikiDiverse}  \\
    \cmidrule{2-5}
    \cmidrule{6-9}
    \cmidrule{10-13}
    ~ & H@1 & H@3 & H@5 & MRR & H@1 & H@3 & H@5 & MRR & H@1 & H@3 & H@5 & MRR \\
    \midrule
    BLINK & 58.47 & 81.51  & 88.09  & 71.39 & 74.66  & 86.63  & 90.57 & 81.72 & 57.14 & 78.04 & 85.32 & 69.15 \\ 
    BERT & 59.55 & 81.12 & 87.16 & 71.67 & 74.82 & 86.79 & 90.47  & 81.78 & 55.77 & 75.73 & 83.11 & 67.38 \\ 
    RoBERTa & 61.34 & 81.56 & 87.15 & 72.80 & 73.75 & 85.85 & 89.80 & 80.86 & 59.46 & 78.54 & 85.08 & 70.52 \\
    \midrule
    CLIP & 67.78 & 85.22 & 90.04 & 77.57 & 83.23 & 92.10 & 94.51 & 88.23 & 61.21 & 79.63 & 85.18 & 71.69\\ 
    ViLT & 45.85 & 62.96 & 69.80 & 56.63 & 72.64 & 84.51 & 87.86 & 79.46 & 34.39 & 51.07 & 57.83 & 45.22\\
    ALBEF & 65.17 & 82.84 & 88.28 & 75.29 & 78.64 & 88.93 & 91.75 & 84.56 & 60.59 & 75.59 & 81.30 & 69.93\\
    METER & 63.96 & 82.24 & 87.08 & 74.15 & 72.46 & 84.41 & 88.17 & 79.49 & 53.14 & 70.93 & 77.59 & 63.71\\
    \midrule
    DZMNED & 68.16 & 82.94 & 87.33 & 76.63 & 78.82 & 90.02 & 92.62 & 84.97 & 56.90 & 75.34 & 81.41 & 67.59\\ 
    JMEL & 48.82 & 66.77 & 73.99 & 60.06 & 64.65 & 79.99 & 84.34 & 73.39 & 37.38 & 54.23 & 61.00 & 48.19\\ 
    VELML & 67.71 & 84.57 & 89.17 & 77.19 & 76.62 & 88.75 & 91.96 & 83.42 & 54.56 & 74.43 & 81.15 & 66.13\\ 
    GHMFC & 72.92 & 86.85 & 90.60 & 80.76 & 76.55 & 88.40 & 92.01 & 83.36 & 60.27 & 79.40 & 84.74 & 70.99\\
    MIMIC & 81.02 & 91.77 & 94.38 & 86.95 & 87.98 & 95.07 & 96.37 & 91.82 & 63.51 & 81.04 & 86.43 & 73.44\\
    \midrule
    ATT & 78.33 & 91.16 & 94.33 & 85.28 & 88.20 & 94.87 & 96.52 & 91.93 & 63.38 & 81.95 & \underline{87.25} & 73.82\\
    OT-MEL (\textit{OT}) & \textbf{83.30} & \textbf{92.39} & \textbf{94.83} & \textbf{88.27} & \textbf{88.97} & \textbf{95.63} & \textbf{96.96} & \textbf{92.59} & \textbf{66.07} & \textbf{82.82} & \textbf{87.39} & \textbf{75.43}\\
    OT-MEL (\textit{KD}) & \underline{82.57} & \underline{92.06} & \underline{94.44} & \underline{87.78} & \underline{88.37} & \underline{95.41} & \underline{96.90} & \underline{92.26} & \underline{64.82} & \underline{82.29} & \underline{87.25} & \underline{74.65}\\
    \bottomrule
    \end{tabular*}
    \caption{Experimental results (\%) on three MEL benchmark datasets, including RichpediaMEL, WikiMEL and WikiDiverse. The \textbf{bold} scores are the best results, and the \underline{underline} scores are the second results.}
    \label{tab:MainResults}
\end{table*}


\section{Experiments}

In this section, we carried out comprehensive experiments on three public MEL benchmark datasets to evaluate the effectiveness of OT-MEL model.

\subsection{Datasets}

In the experiments, we selected three public MEL datasets RichpediaMEL with 17K samples, WikiMEL~\citep{wang2022multimodal} with 25K samples and WikiDiverse~\citep{wang2022wikidiverse} with 15K samples to verify the effectiveness of our proposed method. We follow the training, validation, and testing set splitting approach as \citet{luo2023multi}. Following \citet{wang2022multimodal} we use the subset of Wikidata as KG for each dataset.The detail of three datasets information are reported in Appendix~\ref{sec:dataset_info}.

\subsection{Baselines}
We compare our method with various competitive baselines in three groups:
1) the text-based methods, which utilize textual information only to achieve EL, including \textbf{BLINK}~\citep{wu2019scalable}, \textbf{BERT}~\citep{devlin2018bert}, and \textbf{RoBERTa}~\citep{liu2019roberta}.
2) the vision-and-language pre-training (VLP) methods including \textbf{CLIP}~\citep{radford2021learning}, \textbf{ViLT}~\citep{kim2021vilt}, \textbf{ALBEF}~\citep{li2021align}, and \textbf{METER}~\citep{dou2022empirical}.
3) the MEL enhanced methods\footnote{For fairness, we don't compare with DRIN~\cite{xing2023drin}, since we fail to run the model due to the numerous scene graphs of entities.}, which explore sophisticated interaction networks to capture diverse correlations, including \textbf{DZMNED}~\citep{moon2018multimodal}, \textbf{JMEL}~\citep{adjali2020multimodal}, \textbf{VELML}~\citep{zheng2022visual}, \textbf{GHMFC}~\citep{wang2022multimodal}, \textbf{MIMIC}~\citep{luo2023multi}.
For details, we provide extensive descriptions of baselines in Appendix~\ref{sec:des_baseline}.


\subsection{Implementation Details}

Our model weights are initialized with pre-trained CLIP-Vit-Base-Patch32. 
We set the batch size to 256, the learning rate to $2 \times 10^{-5}$, and the hidden dimension $d$ to 96. For the OT parameter $\lambda$, we set it to 0.6.
All methods are evaluated on the validation set and the checkpoint with the highest MRR is selected to evaluate on the test set.

\subsection{Main Results}

We conduct comparative experiments on three public datasets to verify the effectiveness of each model on MEL tasks, Table~\ref{tab:MainResults} shows the H@k and MRR of different models in the three datasets, and we will provide the formula for metrics in the Appendix~\ref{sec:metrics}. Based on the experimental results, we further have the following observations and analysis:
(1) \textit{{Multimodal information is important for EL problems.}} The first section of the table presents text-based EL models, which, despite demonstrating commendable performance, significantly lag behind the multimodal models showcased in the second section of the table. 
(2) \textit{{Our model is effective on all three datasets.}} Compared with the state-of-the-art model, OT-MEL improves H@1 by 2.28\%, 2.58\% and 0.99\% on the three datasets, respectively. 
(3) \textit{{OT can indeed guide the model for better modality fusion.}} 
The results of OT-MEL(KD) based on distillation on 3 datasets also exceed the current state-of-the-art models.

\subsection{Variant Analysis}

\begin{table}[!t]
    \centering\footnotesize
    \scalebox{0.98}{
    \begin{tabular*}{0.48 \textwidth}{@{\extracolsep{\fill}}@{}l|l|cccc@{}}
    \toprule
    No. & Variant    & H@1 &  MRR & $\Delta$Avg \\
    \midrule
    - & OT-MEL (OT) &   83.30  & 88.27 & - \\
    \midrule
    F1 & repl. ATT   & 80.60  & 86.70 & $\downarrow$ 2.14\\ 
    F2 & w/o   FusM   & 78.94  & 85.42 & $\downarrow$ 3.61\\ 
    \midrule
    M1 & repl. ATT   & 81.72  & 87.61  & $\downarrow$ 1.12\\ 
    M2 & w/o   UniM   & 73.95  & 81.85 & $\downarrow$ 7.89\\ 
    \midrule
    A1 & ATT (w/o OT)   & 78.33  & 85.28 & $\downarrow$ 3.98\\ 
    \midrule
    O1 & SoftPool $\rightarrow$ Mean   & 80.80 & 86.55 & $\downarrow$ 2.11\\ 
    O2 & SoftPool $\rightarrow$ Max   & 82.28  & 87.25 & $\downarrow$ 1.02\\ 
    \midrule
    \midrule
    - & OT-MEL (KL)   & 82.57  & 87.78 & - \\ 
    \midrule
    K1 & KD $\rightarrow$ MSE   & 81.95  & 87.34 & $\downarrow$0.53 \\
    K2 & KD $\rightarrow$ UniKL   & 81.84 & 87.33 & $\downarrow$0.59  \\ 
    K3 & ATT (w/o KD)  & 78.33  & 85.28 & $\downarrow$3.37 \\ 
    \bottomrule
    \end{tabular*}
    }
    \caption{Results (\%) of variants on the Richpediamel data. $\Delta$Avg is the average decrease in MRR and H@1. The details and full results are reported in Appendix~\ref{sec:details_variant_analysis}.}
    \label{tab:VariantAnalysis}
\end{table}

\begin{table*}[t]
    \centering\footnotesize
    \setlength{\tabcolsep}{5pt}
    \begin{tabular*}{\textwidth}{@{\extracolsep{\fill}}@{}l|cccc|cccc|cccc@{}}
    \toprule
    \multirow{2}{*}{Model} &  \multicolumn{4}{c|}{Single-object Image} & \multicolumn{4}{c|}{Multi-object Image} &
    \multicolumn{4}{c}{Overall} \\
    \cmidrule{2-5}
    \cmidrule{6-9}
    \cmidrule{10-13}
    ~ & H@1 & H@3 & H@5 & MRR & H@1 & H@3 & H@5 & MRR & H@1 & H@3 & H@5 & MRR \\
    \midrule
    MIMIC & 75.10 & 87.82 & 91.46 & 82.34 & 79.41 & 92.78 & 95.97 & 86.50 & 75.85 & 88.69 & 92.28 & 83.07 \\ 
    ATT & 76.09 & 90.51 & 93.71 & 83.81 & 75.16 & 91.08 & 95.12 & 83.82 & 71.77 & 88.43 & 92.58 & 80.81 \\
    OT-MEL (\textit{OT}) & \textbf{82.28} & \textbf{92.16} & \textbf{94.87} & \textbf{87.72} & \textbf{85.56} & \textbf{94.69} & 
    \textbf{96.18} & \textbf{90.31} & \textbf{78.82} & \textbf{90.17} & \textbf{93.21} & \textbf{85.02} \\ 
    \bottomrule
    \end{tabular*}
    \caption{Multi-element correlation analysis on Richpediamel according to images with different object numbers.}
    \label{tab:ObjectResults}
\end{table*}

\begin{table}[ht]
    \centering\footnotesize
    \setlength{\tabcolsep}{5pt}
    \scalebox{0.98}{
    \begin{tabular*}{0.48\textwidth}
    {@{\extracolsep{\fill}}@{}c|ccc@{}}
    \toprule
    Model & Richpediamel & Wikimel & Wikidiverse \\
    \midrule
    MIMIC & 1,033.47s & 1,000.12s & 685.69s \\
    ATT & 1,093.03s & 1,875.24s & 1,543.87s \\
    OT-MEL (\textit{KD}) & 1,842.96s & 2,012.95s & 1,595.17s  \\
    OT-MEL (\textit{OT}) & 2,175.66s & 3,028.11s & 1,841.67s \\
    \bottomrule
    \end{tabular*}}
    \caption{Efficiency comparison (s) in prediction.}
    \label{tab:Speed}
\end{table}

To inspect the effectiveness of the model compo- nents, we show results of experiments for model variants in Table~\ref{tab:VariantAnalysis}, where we find:
(1) \textit{{OT is more effective than attention mechanisms in both unimodal and multimodal feature matching.}} In models F1 and M1, where we replaced OT with attention, there was a noticeable decrease in performance for both.
(2) \textit{{Multimodal feature fusion can better utilize the complementary information between modalities to aid in linking.}} In model F2, after removing the multimodal feature fusion module, the average results decreased by 3.61\%.
(3) \textit{{Unimodal feature matching can more effectively utilize the fine-grained correlations between images and texts to enhance EL.}} In model M2, when the unimodal feature matching module is removed, the performance decreases by 7.89\%.
(4) \textit{{Compared to mean pooling and max pooling, SoftPool can aggregate sequential data more effectively.}} After replacing SoftPool with mean pooling and max pooling, the performance decreased by 2.11\% and 1.02\%, respectively.
(5) \textit{{OT-MEL(KL) can enhance the inference speed of the model while only marginally compromising model performance.}} OT-MEL(KL), compared to OT-MEL(OT), improves the inference speed by 337.20s on the validation set of the Richpediamel dataset, while the performance only decreases by 0.62\%. Table~\ref{tab:Speed} displays the average inference speeds on the validation sets of different datasets.

\subsection{Analysis on MEL with Multi-objects}\label{sec:multi-e}

To validate that our model can better utilize the multi-element correlations between different modalities to aid in EL, we used YOLOv5 for object detection on images in the Richpediamel test dataset. 
We categorized images into multi-object sets for object counts of two or more, and single-object sets for counts of one or fewer. Notably, we excluded data where EL was straightforward due to identical mentions and entity names with no ambiguous counterparts in the KG, ensuring test rigor.
Table~\ref{tab:ObjectResults} shows the results of our experiment. 
Observations indicate that \textit{{our model can indeed achieve better linking by utilizing the multi-element correlations between multimodalities}}. Compared to the attention mechanism, our model shows a 6.19\% improvement in H@1 on single-object sets and a 10.40\% improvement in H@1 on multi-object sets.

\subsection{Case Study of Optimal Transport}
\begin{figure}[!t]
    \centering
    \includegraphics[width=0.48\textwidth]{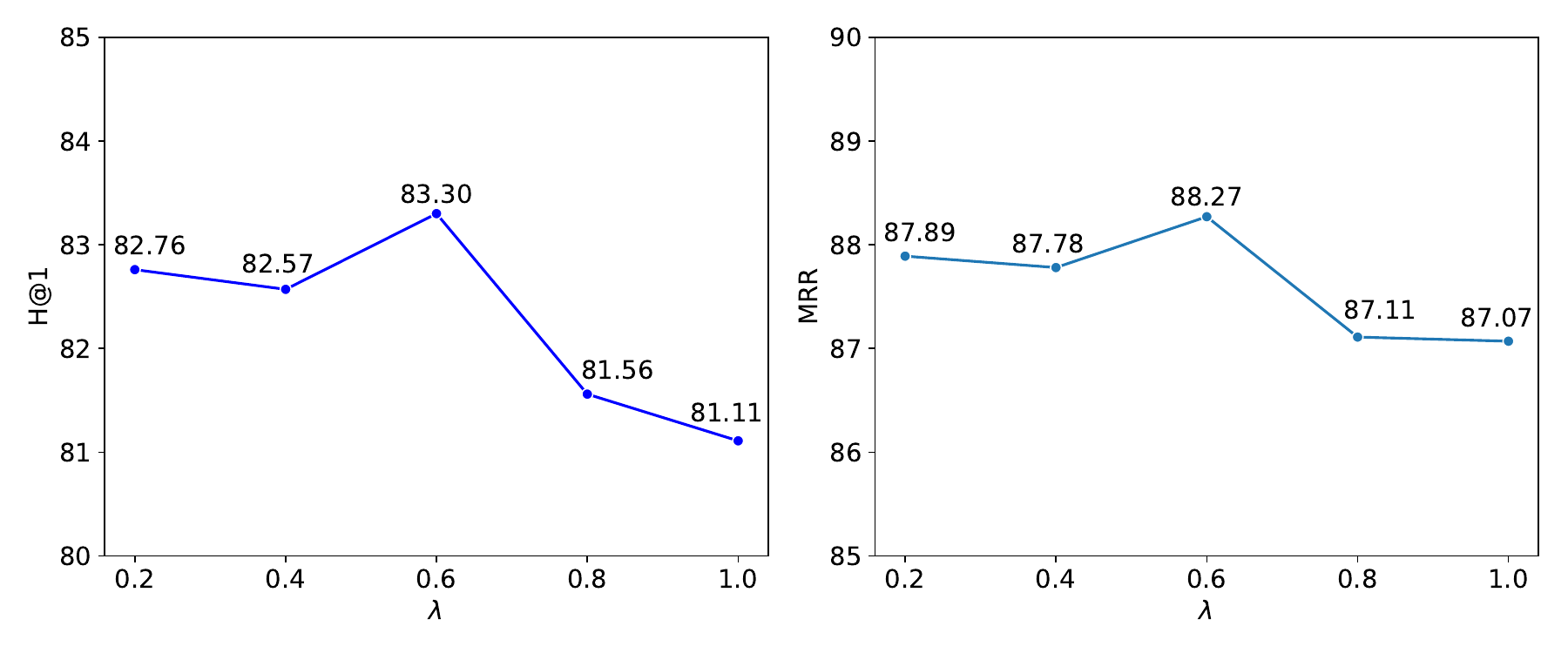}
    \caption{Impact of entropy $H(\bm{\Pi})$ on Richpediamel.}
    \label{fig:entropy}
\end{figure}

The entropy controls the information dispersion in the transport plan of OT, where the higher entropy tends to have a more disperse concentration. 
Figure~\ref{fig:entropy} shows H@1 and MRR results. 
The observation reveals that \textit{{an appropriate entropy allows the model to better focus on the information it should}}. 
We can see that when the entropy is low, the results are poor due to an overly concentrated focus. 
Besides, when the entropy is high, the results tend to be worse due to an overly concentrated focus.

\begin{figure}[!t]
    \centering
    \includegraphics[width=0.48\textwidth]{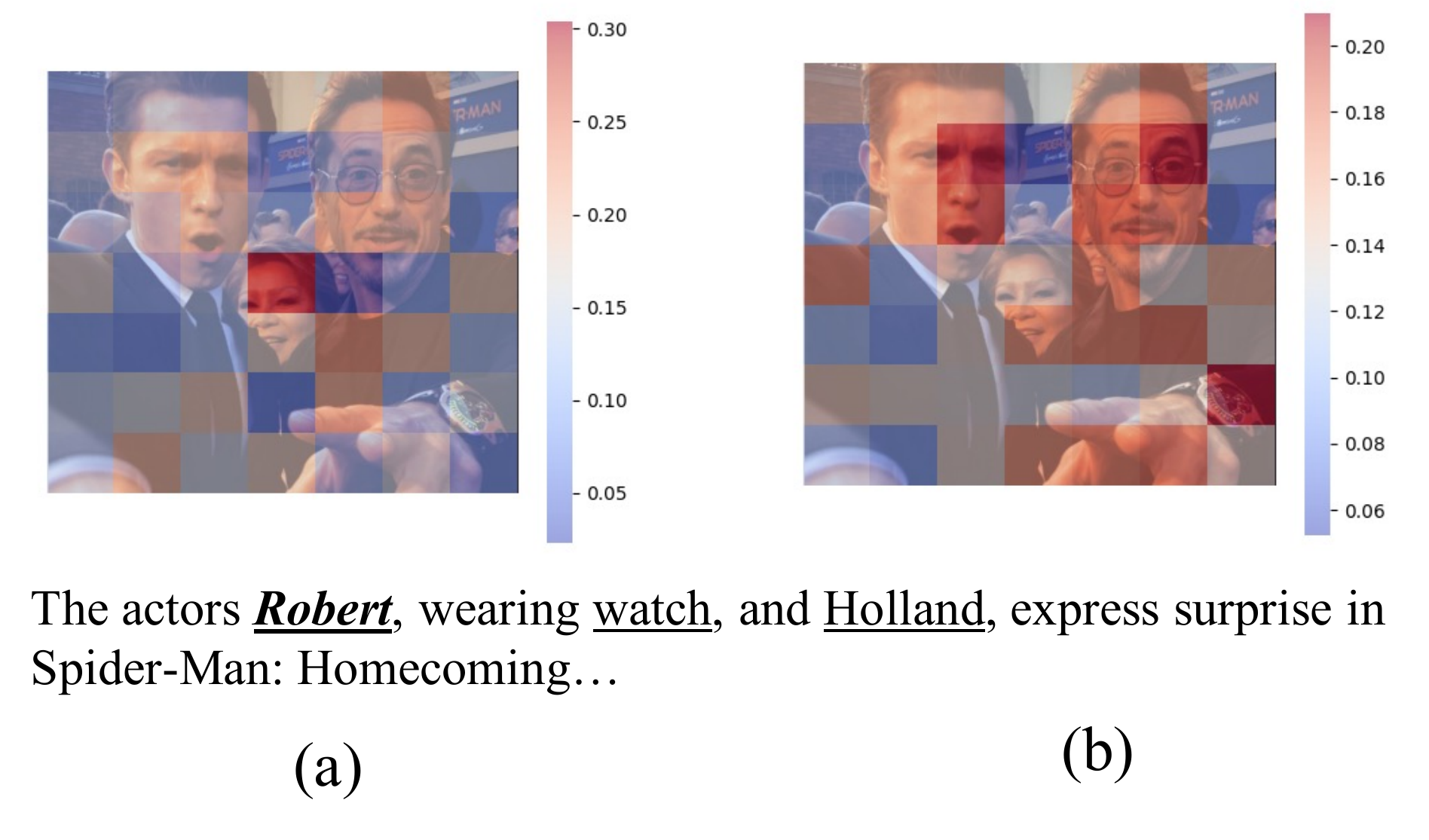}
    \caption{Case study of multi-element correlation assignments: (a) attention matrix, (b) OT matrix.}
\end{figure}


\section{Related Works}

\paragraph{Text-based Entity Linking}
Traditional EL mainly focuses on the textual information of mentions and entities.
These methods~\cite{cao2017bridge, wu2019scalable} first employ textual encoders to obtain contextual representations of mention and entity, and then calculate similarities between them as final probabilities.
To enhance consistency in EL, they enrich local correlations~\cite{peters2019knowledge, liu2023towards} with prior knowledge and collective mechanism, and global correlations~\cite{le2018improving, cao2018neural, fang2019joint} with topical coherence among mentions.
However, these methods are designed to deal with text only, which can hardly deal with the crucial multimodal correlations within images in MEL.
Other studies have focused on Generative Entity Linking (GEL)~\citep{de2020autoregressive, shi2023generative, de2022multilingual, huang2022autoregressive, yuan2022generative}. This is a promising research problem in the era of Large Language Models (LLMs), but they can suffer from the lexical ambiguity problem~\citep{fortuny2024ambiguity}, which we will discuss in details in the Appendix~\ref{sec:gen}.

\paragraph{Multimodal Entity Linking}
Recent years have seen numerous social media and news posted with texts and images, thereby MEL is proposed to identify interested entities from multimodal information, which can benefit to knowledge acquisition~\citep{sheng2020adaptive, li2023attribute}.
As a pioneering study, \citet{moon2018multimodal} firstly introduces images to benefit EL, which blends visual, word and character features through an attention mechanism.
Afterward, there exist several studies~\cite{adjali2020multimodal, zheng2022visual,wang2022multimodal,luo2023multi} exploring gate or attention mechanism to capture complicated multimodal correlations for mentions and entities.
Recently, \citet{xing2023drin} introduce external object-level scene graphs of images to enrich visual correlations, but may involve elaborate errors and still construct multimodal correlations with attentive graph neural networks.
\citet{luo2023multi} present a sophisticated gated framework for fine-grained inter-modal interactions, achieving the most advanced results.
Although existing studies attempt multiple correlations in different aspects, they heavily rely on the automatically learned attention-based mechanism, which can hardly ensure appropriate correlation assignments and impede further improvements.

\paragraph{Optimal Transport} 
OT~\citep{kantorovich1942translocation} is a well-known problem, which can trace back to~\citet{monge1781memoire}.
The key idea is to derive an optimal transport plan to transfer one distributions to another~\citep{arjovsky2017wasserstein, chen2020uniter, bhardwaj2021knot, yang2023multimodal}. There are extensive studies utilizing the transport plan of OT to solve assignment problems~\citep{chang2022unified, cao2022otkge}.
\citet{liu2020semantic} employ the global OT plan to address the many-to-one issue in semantic correspondence in computer vision.
\citet{lee2019hierarchical, chen2020uniter} employ the wasserstein distance to align the representation of text and image.
However, few existing studies conduct comprehensive investigation of the OT plan in the context of multi-element correlations within both unimodal and multimodal settings.

\section{Conclusion}
This paper introduces a MEL model with OT-guided correlation assignment, termed OT-MEL. 
Previous studies rely heavily on the automatically learned attention weights, which may over-concentrate on partial correlations.
To this end, OT-MEL proposes capturing these multi-element correlations across modalities with OT. 
To improve the efficiency, we further transfer knowledge from OT to attention mechanism. 
Experiments on 3 datasets illustrate that our model outperforms current state-of-the-art models. 
In the future, we will explore other efficient method to deploy OT in MEL.

\section*{Limitations}
In this paper, we propose a MEL model with OT correlation assignment, namely OT-MEL.
First, OT-MEL simplistically assumes uniform distributions for both source and target, which can be further explored in real-world scenarios. 
Besides, as OT-MEL is tailored for visual and textual modality in MEL, it doesn't consider OT in other modalities such as speech and video.
We will explore the generality of the proposed OT correlation assignment in these complex modalities for wide research communities.

\section*{Acknowledgement}
We would like to thank the anonymous reviewers for their comments. This work was supported by the National Key Research and Development Program of China (Grant No.2021YFB3100600), the Youth Innovation Promotion Association of CAS (No.2021153), and the Postdoctoral Fellowship Program of CPSF (No.GZC20232968).

\bibliography{anthology,custom}

\clearpage
\appendix

\section{Optimal Transport}
\label{sec:appendix_ot}

\begin{figure}[ht]
    \centering
    \includegraphics[width=\columnwidth]{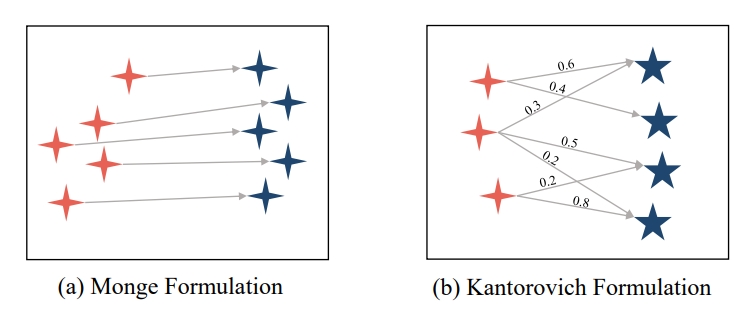}
    \caption{The differences between Monge and Kantorovich formulations in optimal transport, with a multimodal problem example.}
    \label{fig:ot}
\end{figure}

Optimal Transport (OT) is a problem which traces back
to the work of Gaspard Monge~\citep{monge1781memoire}. It is a constrained optimization problem that seeks an efficient solution of transporting one distribution of mass to another. Let $\mu$ and $\nu$ be two probability measures defined on the same space $\mathbb{R}^n$, representing the source and target distributions of resources, respectively. The Monge problem aims to find a mapping $\Pi: \mathbb{R}^n \rightarrow \mathbb{R}^n$ that transforms a distribution $\mu$ into a distribution $\nu$ while minimizing the total transfer cost. This cost is usually measured by a cost function $c(x, \Pi(x))$, where $x$ is the original location of the resource and $\Pi(x)$ is the new location to which it has been transferred. It can be formulated as the following optimization problem:
\begin{equation}\label{eq:Monge}
\Pi^* = \underset{\Pi}{\mathrm{\arg \min}} \int_{\mathbb{R}^n} c(x, \Pi(x)) \, d\mu(x),
\end{equation}
where $\Pi$ is the transport map, $c(x, \Pi(x))$ is a cost function, measure from the $x$ to the $\Pi (x)$ transfer cost, and the integral is the total cost over all $x$.

However, there is a very strong assumption in the Monge problem which is the inseparability of masses, that is, the Monge problem requires an explicit mapping $\Pi$ to transfer directly from one location to another, and when the two distributions have different total masses or there is a many-to-many mapping between the source distribution and the target distribution, the Monge problem cannot find a feasible solution. For example, in multimodal problem, the total mass in the image distribution and the total mass in the text distribution may be different, and an image patch may correspond to multiple text tokens. Kantorovich~\citep{kantorovich1942translocation} proposed an extension and generalization of the Monge problem to address the aforementioned limitations in the Monge problem. Figure~\ref{fig:ot} illustrates the difference between the Monge Formulation and the Kantorovich Formulation.

\begin{algorithm}
\caption{Sinkhorn-Knopp Algorithm}\label{Sinkhorn}
\begin{algorithmic}
\REQUIRE Cost matrix $\mathbf{C} \in \mathbb{R}^{n \times m}$, marginals $\bm{\mu} \in \mathbb{R}^n$, $\bm{\nu} \in \mathbb{R}^m$, regularization parameter $\lambda > 0$
\ENSURE Approximate transport matrix $\mathbf{P}$
\STATE Initialize $\mathbf{u} \gets \mathbf{1}_n$, $\mathbf{v} \gets \mathbf{1}_m$
\STATE Compute $K \gets \exp(-\lambda \mathbf{C})$
\WHILE{not converged}
    \STATE Update $\mathbf{u} \gets \bm{\mu} \oslash (K\mathbf{v})$
    \STATE Update $\mathbf{v} \gets \bm{\nu} \oslash (K^T\mathbf{u})$
\ENDWHILE
\STATE Compute $\mathbf{P} \gets \text{diag}(\mathbf{u}) K \text{diag}(\mathbf{v})$
\RETURN $\mathbf{P}$
\end{algorithmic}
\end{algorithm}

Instead of finding a specific transport map, the Kantorovich problem looks for a transport plan, which is a more flexible notion. In this problem, we consider two given probability distributions $\mu$ and $\nu$, defined on two spaces $X$ and $Y$, respectively. The Kantorovich problem aims to find a joint probability distribution (transport plan) $\Pi$ over $X \times Y$ whose marginal distributions are $\mu$ and $\nu$, respectively, while minimizing the overall transport cost. Mathematically, this problem can be expressed as follows:
\begin{equation}\label{eq:Kantorovich}
\begin{aligned}
    \Pi^\star &= \arg \min_{\Pi} \int_{X \times Y} c(x, y) \Pi(x, y) \, dx \, dy, \\
    \text{s.t.}\ &\int_Y \Pi(x, y) dy = \mu,\ \int_X \Pi(x, y) dx = \nu,
\end{aligned}
\end{equation}
where $\Pi$ is a joint probability measure on $X \times Y$, representing the transition plan, and $c(x, y)$ is a cost function for moving from $x \in X$ to $y \in Y$.

Within the scope of our problem, similar to the previous optimal transport (OT) settings~\citep{arase2023unbalanced, chang2022unified}, we simply assume the tokens of the text and the patches of the image as two different uniform distributions, which means that each token or patch has the same amount of information. Each token or patch is a sample from a discrete probability distribution. So we need to use the Kantorovich problem in discrete form to compute the optimal transport plan between text and image. Specifically, let $X = \{x_1, x_2, \ldots, x_n\}$ be the token set of text and $Y = \{y_1, y_2, \ldots, y_m\}$ be the patch set of image. The discrete form of Kantorovich problem can be formulated as finding a "transport plan" $\Pi$, which is a matrix $n \times m$, where each element $\Pi_{ij}$ represents the amount of information to be transported from text token $x_i$ to image patch $y_j$. This plan needs to minimize the total transportation cost while satisfying constraints on the supply and demand of information. Mathematically, this problem can be formulated as the following linear programming problem:
\begin{equation}\label{eq:Kantorovich-discrete1}
\begin{aligned}
    \Pi^\star &= \underset{\Pi \in \mathbb{R}_{+}^{n \times m}}{\mathrm{\arg \min}} \sum_{i=1}^{m} \sum_{j=1}^{n} c_{ij} \Pi_{ij}, \\
    \text{s.t.} \ &\Pi \mathbf{1}_n = \bm{\mu}, \quad \Pi^{\top} \mathbf{1}_m = \bm{\nu},
\end{aligned}
\end{equation}
where $\Pi^*$ is the optimal transport plan, $\Pi_{ij}$ is the amount of information from $x_i$ to $y_j$ in the optimal transport plan, $\mathbb{R}_{+}^{n \times m}$ denotes the space of $n \times m$ real matrices with all non-negative elements, $c_{ij}$ denotes the unit transport cost from $x_i$ to $y_j$. $\mathbf{1}_n$ is an all-ones column vector of dimension $n$, and $\mathbf{1}_m$ is an all-ones column vector of dimension $m$. Since the computation of the Kantorovich problem involves the resolution of a linear program, its cost is prohibitive when dealing with large-scale data. Therefore,~\citet{cuturi2013sinkhorn} proposes to smooth the kantorovich problem with the entropy regularization term:
\begin{equation}
\begin{aligned}
    \Pi^\star &= \underset{\Pi \in \mathbb{R}_{+}^{n \times m}}{\mathrm{\arg \min}} \sum_{i=1}^{m} \sum_{j=1}^{n} c_{ij} \Pi_{ij} - \frac{1}{\lambda} H(\Pi), \\
    \text{s.t.} \ &\Pi \mathbf{1}_n = \bm{\mu}, \quad \Pi^{\top} \mathbf{1}_m = \bm{\nu},
\end{aligned}
\end{equation}
where $H(\Pi)$ is the entropy of $\Pi$, $\lambda > 0$ is regularization coefficient, then solve the optimal transport problem quickly by Sinkhorn-Knopp algorithm. Algorithm~\ref{Sinkhorn} shows the flow of Sinkhorn-Knopp algorithm, where $\oslash$ is element-wise division.

\section{Description of Baselines}\label{sec:des_baseline}

We compared our method with various competitive baselines.
1) The text-based methods include: 
\begin{itemize}[leftmargin=*]
\item \textbf{BLINK}~\citep{wu2019scalable} a two-stage zero-shot EL method, utilizes BERT as its backbone. It first retrieves entities using a bi-encoder, then re-ranks them with a cross-encoder. 
\item \textbf{BERT}~\citep{devlin2018bert} consists of a stack of Transformer encoders and is pre-trained on a large amount of corpus. 
\item \textbf{RoBERTa}~\citep{liu2019roberta} further improves BERT by removing next sentence prediction objective and using a dynamic mask strategy.
\end{itemize}
2) The Vision-and-Language Pre-training (VLP) methods include: 
\begin{itemize}[leftmargin=*]
\item \textbf{CLIP}~\citep{radford2021learning} uses two Transformer-based encoders for visual and textual representation, pre-trained on massive noisy web data using contrastive loss. 
\item \textbf{ViLT}~\citep{kim2021vilt} adopts shallow embeddings for text and visuals, focusing on deep modality interaction through a series of Transformer layers. 
\item \textbf{ALBEF}~\citep{li2021align} aligns visual and textual features using imagetext contrastive loss first, fusing them via a multimodal Transformer encoder, and employs momentum distillation for enhanced learning from noisy data. 
\item \textbf{METER}~\citep{dou2022empirical} employs co-attention for semantic interplay between modalities, featuring layers with self-attention, cross-attention modules, and a feedforward network.
\end{itemize}
3) The MEL methods include: 
\begin{itemize}[leftmargin=*]
\item \textbf{DZMNED}~\citep{moon2018multimodal} is a trailblazing MEL method, blending visual, word, and character features through an attention mechanism. 
\item \textbf{JMEL}~\citep{adjali2020multimodal} utilizes unigram and bigram embeddings for text analysis, merging these features via concatenation and a dense layer. For fair comparison, its textual encoder is substituted with a pre-trained BERT model. 
\item \textbf{VELML}~\citep{zheng2022visual} employs the VGG-16 network for object-level visual features and replaces its GRU textual encoder with pre-trained BERT, fusing both modalities using an enhanced attention mechanism. 
\item \textbf{GHMFC}~\citep{wang2022multimodal} introduces a hierarchical cross-attention approach for detailed correlation analysis between text and visual features, utilizing contrastive learning for optimization. 
\item \textbf{MIMIC}~\citep{luo2023multi} presents a framework for fine-grained inter-modal interactions, optimized using contrastive learning.
\end{itemize}

\section{Statistics of Datasets}\label{sec:dataset_info}

\begin{figure*}[!t]
    \centering
    \includegraphics[width=1\textwidth]{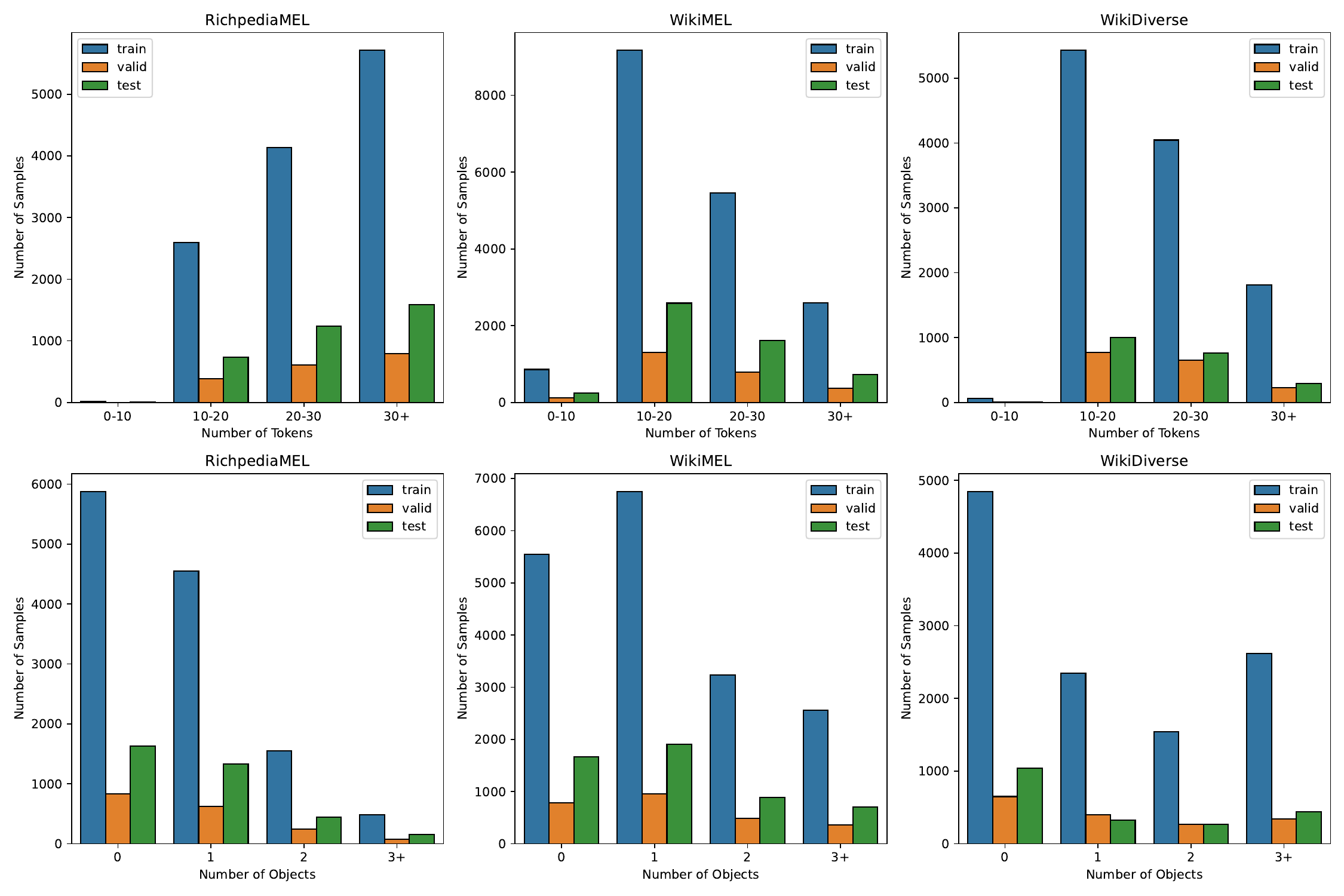}
    \caption{Statistics on text tokens and image objects across three datasets.}
    \label{fig:statistics}
\end{figure*}

\textbf{WikiMEL}~\citep{wang2022multimodal} is collected from Wikipedia entities pages and contains more than 22k multimodal sentences. \textbf{RichpediaMEL}~\citep{wang2022multimodal} is obtained form a MMKG Richpedia~\citep{wang2020richpedia}. The authors of RichpediaMEL first extracted entities form Richpedia and then obtain multimodal information form Wikidata~\citep{wang2022multimodal}. The main entity types of WikiMEL and RichpedaiMEl are person. \textbf{WikiDiverse}~\citep{wang2022wikidiverse} is constructed from Wikinews and covers various topics including sports, technology, economy and so on. Table~\ref{tab:statistics} shows the detail of three datasets. 
Note that we find there also exist some different entities with the same name, and we report the number of them in the Same Name row in Table~\ref{tab:statistics}.
Figure~\ref{fig:statistics} displays statistics on the number of text tokens and the count of objects in images across three datasets.

\begin{table}[ht]
    \centering\footnotesize
    \scalebox{0.98}{
    \begin{tabular*}{0.48 \textwidth}
    {@{\extracolsep{\fill}}@{}c|ccc@{}}
    \toprule
    ~ & Richpediamel & Wikimel & Wikidiverse \\
    \midrule
    Train Mentions & 12,463 & 18,092 & 11,351 \\
    Valid Mentions & 1,780 & 2,585 & 1,664 \\
    Test Mentions & 3,562 & 5,169 & 2,078 \\
    Mention Imgs & 15,853 & 22,136 & 6,697 \\
    \midrule
    Same Name & 15,514 & 616 & 14,827 \\
    KG Entities & 160,933 & 109,976 & 132,460 \\
    Entity with Imgs & 86,769 & 67,195 & 67,309 \\
    \bottomrule
    \end{tabular*}}
    \caption{Statistics of three datasets.}
    \label{tab:statistics}
\end{table}

\begin{figure}[!t]
    \centering
    \includegraphics[width=0.5\textwidth]{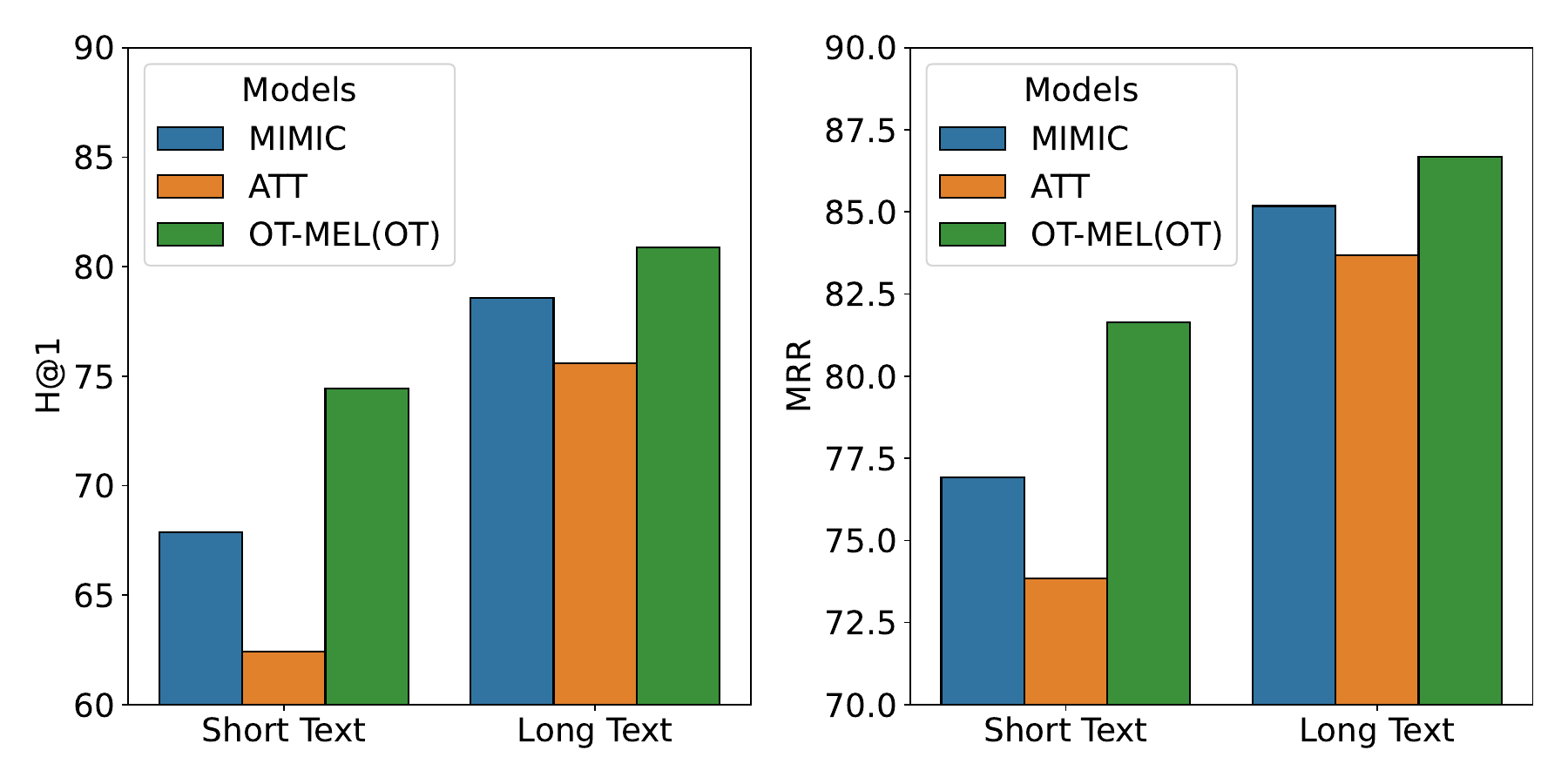}
    \caption{Experiments on samples with different token counts in the Richpediamel dataset.}
    \label{fig:tokenRes}
\end{figure}

\section{Metrics}\label{sec:metrics}

We first calculate the similarity scores between a mention and all entities of the KG, then the similarity scores are sorted in descending order to calculate H@k and MRR, which are defined as:
\begin{equation}
    H@k = \frac{1}{N} \sum_{i}^{N}I(rank(i)<k),
\end{equation}
\begin{equation}
    MRR = \frac{1}{N} \sum_{i}^{N}\frac{1}{rank(i)},
\end{equation}
where $N$ is the number of total samples, $rank(i)$ means the rank of the i-th ground truth entity in the rank list of KG entities, $I(\cdot)$ stands for indicator function which is 1 if the subsequent condition is satisfied otherwise 0.

\section{Details of Variant Analysis}\label{sec:details_variant_analysis}

Table~\ref{tab:Ablation Study} displays all metrics from the variant experiments on the Richpediamel dataset.

\begin{table}[t]
    \centering\footnotesize
    \scalebox{0.98}{
    \begin{tabular*}{0.48 \textwidth}{@{\extracolsep{\fill}}@{}l|l|cccc@{}}
    \toprule
    No. & Variant & H@1 & H@3 & H@5 & MRR \\
    \midrule
    - & OTMEL(OT) & 83.30 & 92.39 & 94.83 & 88.27 \\
    \midrule
    F1 & repl. ATT & 80.60 & 91.66 & 94.50 & 86.70 \\
    F2 & w/o FusM & 78.94 & 90.76 & 93.60 & 85.42 \\
    \midrule
    M1 & repl. ATT & 81.72 & 92.34 & 95.06 & 87.61 \\
    M2 & w/o UniM & 73.95 & 88.12 & 91.86 & 81.85 \\
    \midrule
    A1 & ATT(w/o OT) & 78.33 & 91.16 & 94.33 & 85.28 \\
    \midrule
    O1 & SoftPool $\to$ Mean & 80.80 & 90.90 & 93.91 & 86.55 \\
    O2 & SoftPool $\to$ Max & 82.28 & 92.35 & 94.58 & 87.25 \\
    \midrule
    \midrule
    - & OT-MEL(KL) & 82.57 & 92.06 & 94.44 & 87.78 \\
    \midrule
    K1 & KD $\to$ MSE & 81.95 & 91.66 & 94.22 & 87.34 \\
    K2 & KD $\to$ UniKL & 81.84 & 91.77 & 94.47 & 87.33 \\
    K3 & Att(w/o KD) & 78.33 & 91.16 & 94.33 & 85.28 \\
    \bottomrule
    \end{tabular*}}
    \caption{Results (\%) of variants on Richpediamel data.}
    \label{tab:Ablation Study}
\end{table}

\section{Analysis on Varying Token Counts}

We also conducted extensive experiments on samples with different token counts in the Richpediamel dataset. The experimental results are presented in Figure~\ref{fig:tokenRes}. We categorized texts based on the number of tokens: texts with more than 20 tokens were classified as long texts, while those with fewer than 20 tokens were classified as short texts. Short texts contain fewer linking signals, making the short text group more challenging than the long text group. From our observations, we can draw the following conclusions: \textit{Our model demonstrates stronger robustness in challenging scenarios compared to attention mechanism and baseline models.} In long text scenarios, our model slightly outperforms baseline models and attention mechanisms. However, in the more challenging short text scenarios, while our model only experiences a slight decrease in performance, the baseline models and attention mechanisms show a significant decline.

\section{Dissusion of the Generative Entity Linking}\label{sec:gen}

Generative Entity Linking (GEL)~\citep{de2020autoregressive} utilizes generative models~\cite{lewis2019bart, touvron2023llama} to directly generate the target entity of the mention. Due to the end-to-end paradigm and competitive results, it has received extensive attention~\citep{shi2023generative, de2022multilingual, huang2022autoregressive, yuan2022generative}. 

Generative methods use the constrained decoding mechanism to ensure that the entity names generated by the model must exist in the knowledge graph. 
Specifically, they construct a Trie of alternative entities, where each leaf in the tree is an entity name.
However, they can hardly deal with entities with the same name, which we term as \textbf{lexical ambiguity problem}~\citep{fortuny2024ambiguity}.
For example, \textit{computer scientist Michael Jordan} and \textit{basketball player Michael Jordan} share the same leaf node of the Trie, and we find that there are 14 pages (of entities) in Wikipedia called the same name \textit{Michael Jordan}.

\begin{table}[ht]
    \centering\footnotesize
    \scalebox{0.98}{
    \begin{tabular*}{0.48 \textwidth}
    {@{\extracolsep{\fill}}@{}c|cc@{}}
    \toprule
    Model & Wikimel(Hit@1) & Wikidiverse(Hit@1) \\
    \midrule
    GEMEL & 82.6 & 86.3 \\
    OT-MEL(ours) & 88.97 & 66.07 \\
    \bottomrule
    \end{tabular*}}
    \caption{Comparison with Generative Model.}
    \label{tab:gen}
\end{table}

To further investigate the lexical ambiguity problem, we count the number of these entities with the same name in the KG used for the three MEL datasets in the Table~\ref{tab:statistics}.
We find that the Wikidiverse dataset has more data with lexical ambiguity problem (including 14,827 entities), and the Wikimel dataset has fewer data of that (including 616 entities).
We further compare the results to the typical multimodal generative GEMEL~\citep{shi2023generative}, and show the results in Table~\ref{tab:gen}.
The results demonstrate that, on the \textit{low lexical ambiguity} dataset Wikimel, GEMEL is lower than ours, and better than ours on the \textit{high lexical ambiguity} dataset Wikidiverse.
Although GEMEL performs better on the high lexical ambiguity dataset, it neglects the lexical ambiguity problem, thus cannot distinguish the entities with the same name.

Besides, GENER~\citep{de2020autoregressive}, the unimodal generative method of the EL, can also hardly deal with this problem. 
It tries to solve the lexical ambiguity problem by generating distinguishable entity ids, but the final result drops by about 20\%~\citep{de2020autoregressive}.
Considering the entity ids contain little semantics, how to generate correct ids is challenging for generative models.

With the emergence of LLMs, the GEL is a promising paradigms nowadays. However, how to solve the lexical ambiguity problem of GEL is one of the most important problems in its application, and we will explore it in the future work.

\end{document}